%% file: main.tex
\documentclass[a4paper,conference]{IEEEtran}
\usepackage{graphicx}
\usepackage{amsmath}
\usepackage{amssymb}
\usepackage{rotating}
\usepackage{color}
\usepackage{framed}
\usepackage{multirow}
\usepackage{subcaption}
\usepackage{balance}
\usepackage{booktabs}
\usepackage{url}
\usepackage{bm}
\usepackage{bbm}
\usepackage{arydshln}
\usepackage{xspace}
\usepackage{enumitem}
\usepackage[export]{adjustbox}
\usepackage{lipsum}

\ifCLASSINFOpdf
\else
\fi
\def \ie {\emph{i.e.}}
\def \eg {\emph{e.g.}}
\def \etal {\emph{et al.}}
\newcommand{\tit}[1]{\smallbreak\noindent\textbf{#1.}}

\newcommand{\tinytit}[1]{\noindent\textbf{#1.}}
\def \ours {$\mathsf{eX}^2$\xspace}
\begin{document}
%
\title{\textit{Explore and Explain}: \\Self-supervised Navigation and Recounting}

\author{\IEEEauthorblockN{Roberto Bigazzi, Federico Landi, Marcella Cornia, Silvia Cascianelli, Lorenzo Baraldi, Rita Cucchiara}
\IEEEauthorblockA{University of Modena and Reggio Emilia\\
Email: \{name.surname\}@unimore.it}
}


%


\maketitle

\begin{abstract}
Embodied AI has been recently gaining attention as it aims to foster the development of autonomous and intelligent agents.
In this paper, we devise a novel embodied setting in which an agent needs to explore a previously unknown environment while recounting what it sees during the path. In this context, the agent needs to navigate the environment driven by an exploration goal, select proper moments for description, and output natural language descriptions of relevant objects and scenes. Our model integrates a novel self-supervised exploration module with penalty, and a fully-attentive captioning model for explanation. Also, we investigate different policies for selecting proper moments for explanation, driven by information coming from both the environment and the navigation. Experiments are conducted on photorealistic environments from the Matterport3D dataset and investigate the navigation and explanation capabilities of the agent as well as the role of their interactions.
\end{abstract}


%
\IEEEpeerreviewmaketitle

\input{01-introduction}

\input{02-related}

\input{03-method}

\input{04-experiments}

\input{05-conclusion}

\bibliographystyle{IEEEtran}
\bibliography{bibliography}
%
%
%

\end{document}

%% file: 01-introduction.tex
\section{Introduction}
\label{sec:introduction}
Only a few decades ago, intelligent robots that could autonomously walk and talk existed only in the bright minds of book and movie authors. People used to think about artificial intelligence only as a fictional feature, as the machines they interacted with were purely reactive and showed no form of autonomy. Nowadays, intelligent systems are everywhere, with deep learning being the main engine of the so-called AI revolution.
More recently, advances in the field of embodied AI aim to foster the next generation of autonomous and intelligent robots. Progress in this field includes visual navigation and instruction following~\cite{anderson2018vision}, event though current research is also focused on the creation of new research platforms for simulation and evaluation~\cite{Savva_2019_ICCV,xia2018gibson}. At the same time, tasks at the intersection of computer vision and natural language processing are of particular interest for the community, with image captioning being one of the most active areas~\cite{karpathy2015deep,anderson2018bottom,cornia2020m2}. By describing the content of an image or a video, captioning models can bridge the gap between the black-box architecture and the user.

\begin{figure}[t]
    \centering
    \includegraphics[width=.92\linewidth]{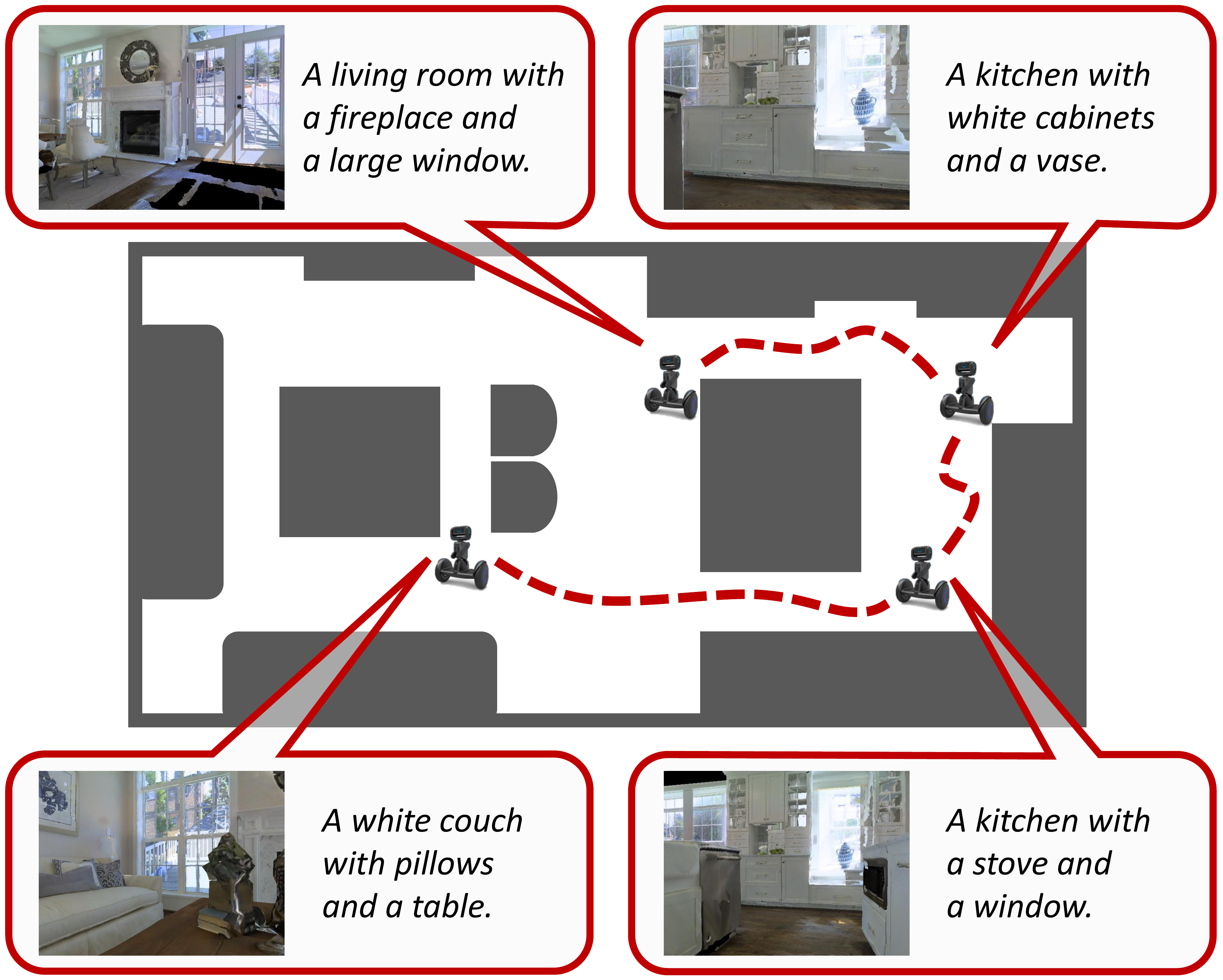}
    \caption{We propose a novel setting in which an embodied agent performs joint curiosity-driven exploration and explanation in unseen environments. While navigating the environment, the agent must produce informative descriptions of what it sees, providing a means of interpreting its internal state.}
    \label{fig:first_page}
    \vspace{-0.15cm}
\end{figure}

In this paper, we propose a new task at the intersection of embodied AI, computer vision, and natural language processing, and aim to create a robot that can navigate through a new environment and describe what it sees.
We call this new task \textit{Explore and Explain} since it tackles the problem of joint exploration and captioning (Fig.~\ref{fig:first_page}). In this schema, the agent needs to perceive the environment around itself, navigate it driven by an exploratory goal, and describe salient objects and scenes in natural language. Beyond navigating the environment and translating visual cues in natural language, the agent also needs to identify appropriate moments to perform the explanation step. 

It is worthwhile to mention that both exploration and explanation feature significant challenges. 
Effective exploration without any previous knowledge of the environment can not exploit a reference trajectory and the agent cannot be trained with classic methods from reinforcement learning~\cite{wijmans2019dd}.
%
%
%
%
To overcome this problem, we design a self-supervised exploration module that is driven solely by curiosity towards the new environment. In this setting, rewards are more sparse than in traditional setups and encourage the agent to explore new places and to interact with the environment.

While we are motivated by recent works incorporating curiosity in Atari and other exploration games~\cite{agrawal2016learning,pathak2017curiosity,burda2018large}, the effectiveness of a curiosity-based approach in a photorealistic, indoor environment has not been tested extensively. Some preliminary studies~\cite{ramakrishnan2020exploration} suggest that curiosity struggles with embodied exploration. In this work, we show that a simple modification of the reward function can lead to striking improvements in the exploration of unseen environments.

Additionally, we encourage the agent to produce a description of what it sees throughout the navigation. In this way, we match the agent internal state (the measure of curiosity) with the variety and the relevance of the generated captions. Such matching offers a proxy for the desirable by-product of interpretability. In fact, by looking at the caption produced, the user can more easily interpret the navigation and perception capabilities of the agent, and the motivations of the actions it takes~\cite{cornia2019smart}. In this sense, our work is related to goal-driven explainable AI, \ie~the ability of autonomous agents to explain their actions and the reasons leading to their decisions~\cite{anjomshoae2019explainable}. 

Previous work on image captioning has mainly focused on recurrent neural networks. However, the rise of Transformer~\cite{vaswani2017attention} and the great effectiveness shown by the use of self-attention have motivated a shift towards recurrent-free architectures. Our captioning algorithm builds upon recent findings on the importance of fully-attentive networks for image captioning and incorporates self-attention both during the encoding of the image features and in the decoding phase. This also allows for a reduction in computational requirements.

Finally, to bridge exploration and recounting, our model can count on a novel speaker policy, which regulates the speaking rate of our captioner using information coming from the agent perception. We call our architecture \ours, from the name of the task: \textit{\textbf{Ex}plore and \textbf{Ex}plain}.

Our main contributions are as follows:
\begin{itemize}
    \item We propose a new setting for embodied AI, \textit{Explore and Explain} in which the agent must jointly deal with two challenging tasks: exploration and captioning of unseen environments.
    \item We devise a novel solution involving curiosity for exploration. Thanks to curiosity, we can learn an efficient policy which can easily generalize to unseen environments.
    \item We are the first, to the best of our knowledge, to apply a captioning algorithm exclusively to indoor environment for robotic exploration. Results are encouraging and motivate further research.
\end{itemize}

%% file: 02-related.tex
\section{Related Work}
\label{sec:related}
Our work is related to the literature on embodied visual exploration, curiosity-driven exploration, and captioning. In the following, we provide an overview of the most important work in these settings, and we briefly describe the most commonly used interactive environments for navigation agents. 

\tit{Embodied visual exploration}
Current research on embodied AI is mainly focused on tasks that require navigating indoor locations. Vision-and-language navigation~\cite{anderson2018vision}, point-goal and object-goal navigation~\cite{wijmans2019dd,anderson2018evaluation,zhu2017target} are all tasks involving the ability for the agent to move across a previously unknown environment. Very recently, Ramakrishnan~\etal~\cite{ramakrishnan2020exploration} highlighted the importance of visual exploration in order to pre-train a generic embodied agent. While their study is mainly focused on exploration as a mean to gather information and to prepare for future tasks, we investigate the role of surprisal for exploration and the consistency between navigation paths and the descriptions given by the agent during the episodes.

\tit{Curiosity-driven exploration}
Curiosity-driven exploration is an important topic in reinforcement learning literature. In this context,~\cite{oudeyer2009intrinsic} provides a good summary on early works on intrinsic motivation. Among them, Schmidhuber~\cite{schmidhuber2010formal} and Sun~\etal~\cite{sun2011planning} proposed to use information gain and compression as intrinsic rewards, while Klyubin~\etal~\cite{klyubin2005empowerment}, and Mohamed and Rezende~\cite{mohamed2015variational} adopted the concept of empowerment as reward during training. 
Differently, Houthooft~\etal~\cite{houthooft2016vime} presented an exploration strategy based on the maximization of information gain about the agent’s belief of environment dynamics. 
Another common approach for exploration is that of using state visitation counts as intrinsic rewards~\cite{bellemare2016unifying,tang2017exploration}. Our work follows the strategy of jointly training forward and backward models for learning a feature space, which has demonstrated to be effective in curiosity-driven exploration in Atari and other exploration games~\cite{agrawal2016learning,pathak2017curiosity,burda2018large}. To the best of our knowledge, we are the first to investigate this type of exploration algorithms in photorealistic indoor environments.

\tit{Interactive environments}
When it comes to the training of intelligent agents, an important role is played by the underlying environment. A first test bed for research in reinforcement learning has been provided by the Atari games~\cite{bellemare2013arcade,brockman2016openai}. However, these kind of settings are not suitable for navigation and exploration in general. To solve this problem, many maze-like environments have been proposed~\cite{kempka2016vizdoom,beattie2016deepmind}. However, agents trained on synthetic environments hardly adapt to real world scenarios, because of the drastic change in terms of appearances.
Simulating platforms like Habitat~\cite{Savva_2019_ICCV}, Gibson~\cite{xia2018gibson}, and Matterport3D simulator~\cite{anderson2018vision} provide a photorealistic environment to train navigation agents. Some of these simulators only provide RGB equirectangular images as visual input~\cite{anderson2018vision}, while others employ the full 3D model and implement physic interactions with the environment~\cite{Savva_2019_ICCV,xia2018gibson}.

\tit{Automatic captioning}
In the last few years, a large number of models has been proposed for image captioning~\cite{anderson2018bottom,xu2015show,rennie2017self}. The majority of them use recurrent neural networks as language models and a representation of the image which might be given by the output of a CNN~\cite{rennie2017self,vinyals2017show}, or by a time-varying vector extracted with attention mechanisms over either a spatial grid of CNN features~\cite{xu2015show} or multiple image region vectors extracted from a pre-trained object detector~\cite{anderson2018bottom}. Regarding the training strategies, notable advances have been made by using reinforcement learning to train non-differentiable captioning metrics~\cite{rennie2017self}.
Recently, following the strong advent of fully-attentive mechanisms in sequence modeling tasks~\cite{vaswani2017attention}, different Transformer-based captioning models have been presented~\cite{cornia2020m2,herdade2019image}. In this work, we devise a captioning model based on the Transformer architecture that, for the first time, is applied to images taken from indoor environments for robotic exploration. 

%% file: 03-method.tex
\begin{figure*}[t]
    \centering
    \includegraphics[width=0.98\linewidth]{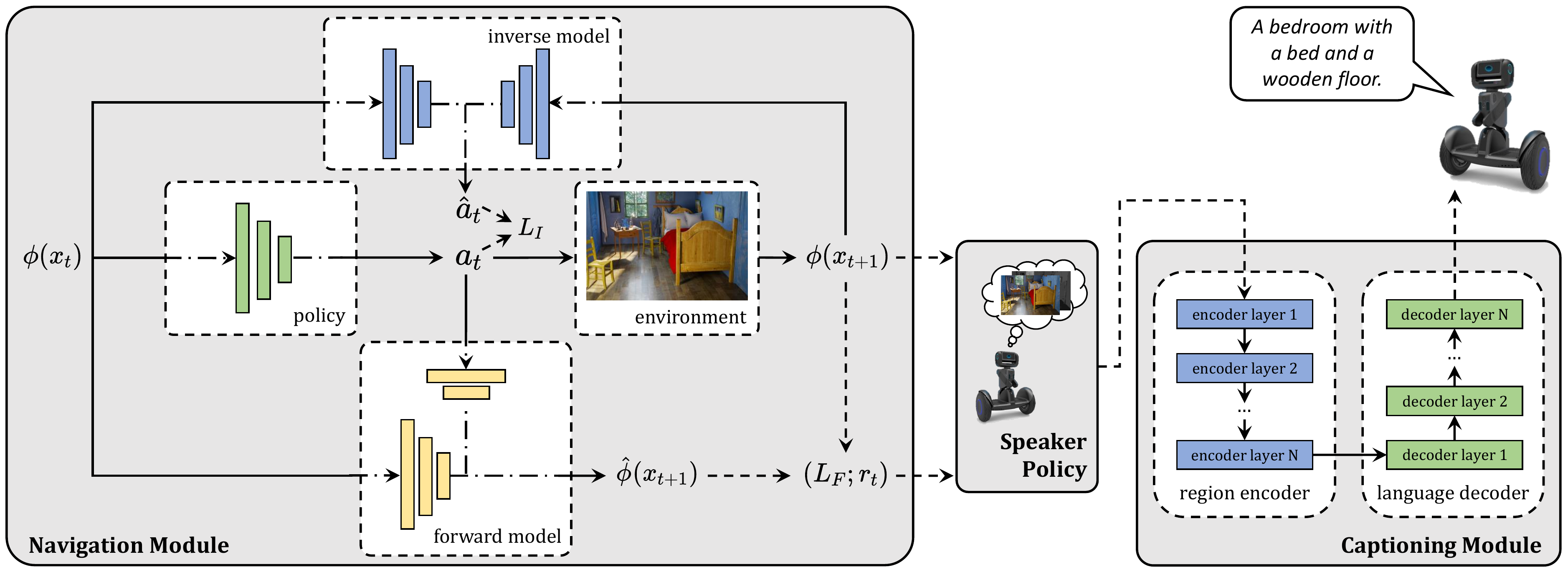}
    \caption{Overview of our \ours framework for navigation and captioning. Our model is composed of three main components: a navigation module which is in charge of exploring the environment, a captioning module which produces a textual sentence describing the agent point of view, and a speaker policy that connects the previous modules and activates the captioning component based on the information collected during the navigation.}
    \label{fig:method}
    \vspace{-0.15cm}
\end{figure*}

\section{Proposed Method}
\label{sec:method}
The proposed method consists of three main parts: a navigation module, a speaker policy, and a captioner. The last two components constitute the speaker module, which is used to explain the agent first-person point of view. The explanation is elicited by our speaker module basing on the information gathered during the navigation. Our architecture is depicted in Fig.~\ref{fig:method} and detailed below.

\subsection{Navigation module}
\label{sec:method_nav}
The navigation policy takes care of the agent displacement inside the environment. At each time step $t$ the agent acquires an observation $x_t$ from the surroundings, performs an action $a_t$, and gets the consequent observation $x_{t+1}$. The moves available to the agent are simple, atomic actions such as \textit{rotate 15 degrees} and \textit{step ahead}.
Our navigation module consists of three main components: a feature embedding network, a forward model, and an inverse model. The discrepancy of the predictions of dynamics models with the actual observation is measured by a reward signal $r_t$, which is then used to stimulate the agent to move towards more informative states.

\tit{Embedding network}
At each time step $t$, the agent observes the environment and gathers $x_t$. This observation corresponds to the raw RGB-D pixels coming from the forward-facing camera of the agent. Yet, raw pixels are not optimal to encode the visual information~\cite{burda2018large}. For this reason, we employ a convolutional neural network $\phi$ to encode a more efficient and compact representation of the surrounding environment. We call this embedded representation $\phi(x_t)$. To ensure that the features observed by the agent are stable throughout the training, we do not change the set of parameters $\theta_\phi$ during training. This approach is shown to be efficient for generic curiosity-based agents~\cite{burda2018large}.


\tit{Forward dynamics model}
Given an agent with policy $\pi(\phi(x_t); \theta_\pi)$, represented by a neural network with parameters $\theta_\pi$, the selected action at timestep $t$ is given by:
\begin{equation}
    a_t \sim \pi \Big(\phi(x_t); \theta_\pi \Big).
\end{equation}
After executing the chosen action, the agent observes a new visual stimulus $\phi(x_{t+1})$. The problem of predicting the next observation given the current input and action to be performed can be defined as a forward dynamics problem:
\begin{equation}
    \hat\phi(x_{t+1}) = f\Big(\phi(x_t), a_t; \theta_F \Big),
\end{equation}
where $\hat\phi(x_{t+1})$ is the predicted visual embedding for the next observation $x_{t+1}$ and $f$ is the forward dynamics model with parameters $\theta_F$.
The forward model is trained to minimize the following loss function:
\begin{equation}
    L_F
    = \frac{1}{2} \left\| \hat\phi(x_{t+1}) - \phi(x_{t+1}) \right\|^2_2 
\label{eq:forward_loss}
\end{equation}

\tit{Inverse dynamics model}
Given two consecutive observations $(x_t, x_{t+1})$, the inverse dynamics model aims to predict the action performed at timestep $t$:
\begin{equation}
    \hat a_t = g \Big(\phi(x_t), \phi(x_{t+1}); \theta_I \Big),
\end{equation}
where $\hat a_t$ is the predicted estimate for the action $a_t$ and $g$ is the inverse dynamics model with parameters $\theta_I$. In our work, the inverse model $g$ predicts a probability distribution over the possible actions and it is optimized to minimize the cross-entropy loss with the ground-truth action $a_t$ performed in the previous time step:
\begin{equation}
    L_I = {y_t \log \hat a_t},
\label{eq:inverse_loss}
\end{equation}
where $y_t$ is the one-hot representation for $a_t$.

\tit{Curiosity-driven exploration}
The agent exploration policy $\pi(\phi(x_t); \theta_\pi)$ is trained to maximize the expected sum of
rewards:
\begin{equation}
    \max_{\theta_\pi} \mathbb{E}_{\pi (\phi(x_t); \theta_\pi)} \left[ \sum_t r_t \right],
\label{eq:reward}
\end{equation}
where the exploration reward $r_t$ at timestep $t$, also called surprisal~\cite{achiam2017surprise}, is given by our forward dynamics model:
\begin{equation}
    r_t = \frac{\eta}{2} \left\| f\big(\phi(x_t), a_t \big) - \phi(x_{t+1}) \right\|^2_2 ,
\label{eq:surprisal}
\end{equation}
with $\eta$ being a scaling factor.
The overall optimization problem can be written as a composition of Eq.~\ref{eq:forward_loss},~\ref{eq:inverse_loss}, and~\ref{eq:reward}:
\begin{equation}
    \min_{\theta_\pi, \theta_F, \theta_I} \bigg[
        - \lambda \mathbb{E}_{\pi (\phi(x_t); \theta_\pi)} \Big[ \mathbf{\Sigma}_t r_t \Big]
        + \beta L_F 
        + (1 - \beta) L_I
    \bigg]
\label{eq:optimization}
\end{equation}
where $\lambda$ weights the importance of the intrinsic reward signal \textit{w.r.t.} the policy loss, and $\beta$ balances the contributions of the forward and inverse models. 

\tit{Penalty for repeated actions}
To encourage diversity in our policy, we devise a  penalty which triggers after the agent has performed the same move for $\tilde t$ timesteps. This prevents the agent from always picking the same action and encourages the exploration of different combinations of atomic actions.

We can thus rewrite the surprisal in Eq.~\ref{eq:surprisal} as:
\begin{equation}
    r_t = \frac{\eta}{2} \left\| f\big(\phi(x_t), a_t \big) - \phi(x_{t+1}) \right\|^2_2 - p_t ,
\label{eq:final_reward}
\end{equation}
where $p_t$ is the penalty at time step $t$. In the simplest formulation, $p_t$ can be modeled with a scalar which is either $0$ or equal to a constant $\tilde p$, after an action has been repeated $\tilde t$ times.

\subsection{Speaker policy}
\label{sec:policy}
As the navigation proceeds, new observations $x_t$ are acquired and rewards $r_t$ are obtained at each time step. Based on these, a speaker policy can be defined, that activates the captioning module. Different types of information from the environment and the navigation module allow defining different policies. In this work, we consider three policies, namely: object-driven, depth-driven, and curiosity-driven. 
\tit{Object-driven policy}
Given the RGB component of the observation $x_t$, relevant objects can be recognized. When at least a minimum number $\mathbf{O}$ of such objects are observed, the speaker policy triggers the captioner. The idea behind this policy is to let the captioner describe the scene only when objects that allow connoting the different views are present.
\tit{Depth-driven policy}
Given the depth component of the observation $x_t$, the speaker policy activates the captioner when the mean depth value perceived $\mathbf{D}$ is above a certain threshold. This way, the captioner is triggered only depending on the distance of the agent from generic objects, regardless of their semantic category.
\tit{Curiosity-driven policy}
Given the surprisal reward defined as in Eq.~\ref{eq:surprisal} and possibly cumulated over multiple timesteps, $\mathbf{S}$, the speaker policy triggers the captioner when $\mathbf{S}$ is above a certain threshold. This policy is independent of the type of information perceived from the environment but is instead closely related to the navigation module. Thus, it helps to match the agent's internal state with the generated captions more explicitly than the other policies.

\subsection{Captioning module}
When the speaker policy activates, a captioning module is in charge of producing a description in natural language given the current observation $x_t$. Following recent literature on the topic, we here employ a visual encoder based on image regions~\cite{ren2017faster}, and a decoder which models the probability of generating one word given previously generated ones. In contrast to previous captioning approaches based on recurrent networks, we propose to use a fully-attentive model for both the encoding and the decoding stage, building on the Transformer model~\cite{vaswani2017attention}.

\tit{Region encoder}
Given a set of features from image regions $R = \{r_1, ..., r_N\}$ extracted from the agent visual view, our encoder applies a stack of self-attentive and linear projection operations. As the former be seen as convolutions on a graph, the role of the encoder can also be interpreted as that of learning visual relationships between image regions. The self-attention operator $S$ builds upon three linear projections of the input set, which are treated as queries, keys and values for an attention distribution. Stacking region features $R$ in matrix form, the operator can be defined as follows: 

\begin{align}
    S(R) = \mathsf{Attention}(W_q R, W_k R, W_v R),\\ \nonumber
\mathsf{Attention}(Q, K, V)=\operatorname{softmax}\left(\frac{Q K^{T}}{\sqrt{d}}\right) V.
\label{eq:attention}
\end{align}
The output of the self-attention operator is a new set of elements $S(R)$, with the same cardinality as $R$, in which each element of $R$ is replaced with a weighted sum of the values, \ie~of linear projections of the input. 


Following the structure of the Transformer model, the self-attention operator $S$ is followed by a position-wise feed-forward layer, and each of these two operators is encapsulated within a residual connection and a layer norm operation. Multiple layers of this kind are then applied in a stack fashion to obtain the final encoder.

\tit{Language decoder}
The output of the encoder module is a set of region encodings $\tilde{R}$ with the same cardinality of $R$. We employ a fully-attentive decoder which is conditioned on both previously generated words and region encodings, and is in charge of generating the next tokens of the output caption. The structure of our decoder follows that of the Transformer~\cite{vaswani2017attention}, and thus relies on self-attentive and cross-attentive operations.

Given a partially decoded sequence of words $W = \{w_0, w_1, ..., w_\tau\}$, each represented as a one-hot vector, the decoder applies a self-attention operation in which $W$ is used to build queries, keys and values. To ensure the causality of this sequence encoding process, we purposely mask the attention operator so that each word can only be conditioned to its left-hand sub-sequence, \ie~word $w_t$ is conditioned on $\{w_{t'}\}_{t' \leq t}$ only. 
Afterwards, a cross-attention operator is applied between $W$ and $\tilde{R}$ to condition words on regions, as follows:
\begin{align}
    C(W, \tilde{R}) = \mathsf{Attention}(W_q W, W_k \tilde{R}, W_v \tilde{R}).
\end{align}

As in the Transformer model, after a self-attention and a cross-attention stage, a position-wise feed-forward layer is applied, and each of these operators is encapsulated within a residual connection and a layer norm operation. Finally, our decoder stacks together multiple decoder layers, helping to refine the understanding of the textual input.

Overall, the decoder takes as input word vectors, and the $t$-th element of its output sequence encodes the prediction of a word at time $t+1$, conditioned on $\{w_t\}_{\le t}$. After taking a linear projection and a softmax operation, this encodes a probability over words in the dictionary. During training, the model is trained to predict the next token given previous ground-truth words; during decoding, we iteratively sample a predicted word from the output distribution and feed it back to the model to decode the next one, until the end of the sequence is reached. Following the usual practice in image captioning literature, the model is trained to predict an end-of-sequence token to signal the end of the caption.


%% file: 04-experiments.tex
\section{Experimental Setup}
\label{sec:experiments}

\subsection{Dataset}
\label{matterport}
The main testbed for this work is Matterport3D~\cite{Matterport3D}, a photorealistic dataset of indoor environments. Some of the buildings in the dataset contain outdoor components like swimming pools or gardens, raising the difficulty of the exploration task. The dataset is split into $61$ scenes for
training, $11$ for validation, and $18$ for testing. It also provides instance segmentation annotations that we use to evaluate the captioning module. Overall, the dataset is annotated with $40$ different semantic categories. For both training and testing, we use the episodes provided by Habitat API \cite{Savva_2019_ICCV} for the point goal navigation task, employing only the starting point of each episode. The size of the training set amounts to a total of $5$M episodes, while the test set is composed of $1\,008$ episodes.

\subsection{Evaluation protocol}
\label{sec:protocol}
\tinytit{Navigation module}
To quantitatively evaluate the navigation module, we use a curiosity-based metric: we extract the sum of the surprisal values defined in Eq.~\ref{eq:surprisal} every $20$ steps performed by the agent, and then we compute the average over the number of test episodes.

\tit{Captioning module}
Standard captioning methods are usually evaluated by comparing each generated caption against the corresponding ground-truth sentences. However, in this setting, only the information on which objects are present on the scene is available, thanks to the semantic annotations provided by the Matterport3D dataset. Therefore, to evaluate the performance of our captioning module, we define two different metrics: a soft coverage measure that assesses how the predicted caption covers all the ground-truth objects, and a diversity score that measures the diversity in terms of described objects of two consecutively generated captions.

In details, for each caption generated according to the speaker policy, we compute the soft coverage measure between the ground-truth set of semantic categories and the set of nouns in the caption. Given a predicted caption, we firstly extract all nouns $\bm{n}$ from the sentence and we compute the optimal assignment between them and the set of ground-truth categories $\bm{c}^*$, using distances between word vectors and the Hungarian algorithm~\cite{kuhn1955hungarian}. We then define an intersection score between the two sets as the sum of assignment profits. Our coverage measure is computed as the ratio of the intersection score and the number of ground-truth semantic classes:
\begin{equation}
    \mathsf{Cov}(\bm{n}, \bm{c}^*) = \frac{\text{I}(\bm{n}, \bm{c}^*)}{\#\bm{c}^*},
\end{equation}
where $\text{I}(\cdot, \cdot)$ is the intersection score, and the $\#$ operator represents the cardinality of the set of ground-truth categories.

Since images may contain small objects which not necessarily should be mentioned in a caption describing the overall scene, we define a variant of the coverage measure by thresholding over the minimum object area. In this case, we consider $\bm{c}^*$ as the set of objects whose overall areas are greater than the threshold.

For the diversity measure, we consider the sets of nouns extracted from two consecutively generated captions, indicated as $\bm{n}_t$ and $\bm{n}_{t+1}$, and we define a soft intersection over union score between the two sets of nouns. Also in this case, we compute the intersection score between the two sets using word distances and the Hungarian algorithm to find the optimal assignment. Recalling
that set union can be expressed in function of an intersection, the final diversity score is computed by subtracting the intersection over union score from $1$ (\ie~the Jaccard distance between the two sets):
\begin{equation}
    \mathsf{Div}(\bm{n}_t, \bm{n}_{t+1}) = 1 - \frac{\text{I}(\bm{n}_t, \bm{n}_{t+1})}{\#\bm{n}_t + \#\bm{n}_{t+1} - \text{I}(\bm{n}_t, \bm{n}_{t+1})},
\end{equation}
where $\text{I}(\cdot, \cdot)$ is the intersection score previously defined, and the $\#$ operator represents the cardinality of the sets of nouns.

We evaluate the diversity of generated captions with respect to the three speaker policies described in Sec.~\ref{sec:policy} and considering different thresholds for each policy (\ie~number of objects, mean depth value, and surprisal score). For each speaker policy and selected threshold, the agent is triggered a different number of times thus generating a variable number of captions during the episode. We define the agent's overall loquacity as the number of times it is activated by the speaker policy according to a given threshold. In the experiments, we report the loquacity values averaged over the test episodes.

\subsection{Implementation and training details}
\tinytit{Navigation module}
Navigation agents are trained using only visual inputs, with each observation converted to grayscale, cropped and re-scaled to a $84 \times 84$ size. A stack of four historical observations $[x_{t-3}, x_{t-2}, x_{t-1}, x_t]$ is used for training in order to model temporal dependencies. We adopt PPO~\cite{schulman2017proximal} as learning algorithm and employ Adam~\cite{kingma2015adam} as optimizer. The learning rate for all networks is set to $10^{-4}$ and the length of rollouts is equal to $128$. For each rollout we make 3 optimization epochs.
The features $\phi(x_t)$ used by the forward and backward dynamics networks are $512$-dimensional and are obtained using a randomly initialized convolutional network $\phi$ with fixed weights $\theta_\phi$, following the approach in~\cite{burda2018large}.

The model is trained using the splits described in Sec.~\ref{matterport}, stopping the training after $10\,000$ updates of the agent. The length of an exploration episode is $1\,000$ steps. In our experiments, we set the parameters reported in Eq.~\ref{eq:optimization} to $\lambda=0.1$ and $\beta=0.2$, respectively.
Concerning the penalty $p_t$ given to the agent to stimulate diversity (Eq.~\ref{eq:final_reward}), we set $p_t = \tilde p = 0.01$ after the same action is repeated for $\tilde{t}=5$ times. 

\tit{Speaker policy}
For the object-driven policy, we use the instance segmentation annotations provided by the Matterport3D simulator. For this policy, we select $15$ of the $40$ semantic categories in the dataset, discarding the contextual ones, which would not be discriminative for the different views acquired by the agent, as for example \textit{wall}, \textit{floor}, and \textit{ceiling}. This way, we can better evaluate the effect of the policy without it being affected by the performance of an underlying object detector of recognizing objects in the agent's current view.
Also for the depth-driven policy, we obtain the depth information of the current view from the Matterport3D simulator, averaging the depth values to extract a single score. 
In the curiosity-driven policy, we consider the sum of surprisal scores extracted over the last 20 steps, obtained by the agent during navigation.

\tit{Captioning module}
To represent image regions, we use Faster R-CNN~\cite{ren2017faster} finetuned on the Visual Genome dataset~\cite{krishnavisualgenome,anderson2018bottom}, thus obtaining a $2048$-dimensional feature vector for each region. To represent words, we use one-hot vectors and linearly project them to the input dimensionality of the model, $d$. We also employ sinusoidal positional encodings~\cite{vaswani2017attention} to represent word positions inside the sequence, and sum the two embeddings before the first encoding layer. 
In both region encoder and language decoder, we set the dimensionality $d$ of each layer to $512$, the number of heads to $8$, and the dimensionality of the inner feed-forward layer to $2048$. We use dropout with keep probability $0.9$ after each attention layer and after position-wise feed-forward layers. 

Following a standard practice in image captioning~\cite{rennie2017self,anderson2018bottom}, we train our model in two phases using image-caption pairs coming from the COCO dataset~\cite{lin2014microsoft}. Firstly, the model is trained with cross-entropy loss to predict the next token given previous ground-truth words. Then, we further optimize the sequence generation using reinforcement learning employing a variant of the self-critical sequence training~\cite{rennie2017self} on sequences sampled using beam search~\cite{anderson2018bottom}. 
Pre-training with cross-entropy loss is done using the learning rate scheduling strategy defined in~\cite{vaswani2017attention} with a warmup equal to $10\,000$ iterations. Then, during finetuning with reinforcement learning, we use the CIDEr-D score~\cite{vedantam2015cider} as reward and a fixed learning rate equal to $5^{-6}$. We train the model using the Adam optimizer~\cite{kingma2015adam} and a batch size of $50$. During CIDEr-D optimization and caption decoding, we use beam search with a beam size equal to $5$. 
To compute coverage and diversity metrics and for extracting nouns from predicted captions, we use the spaCy NLP toolkit\footnote{\url{https://spacy.io/}}. We use GloVe word embeddings~\cite{pennington2014glove} to compute word similarities between nouns and semantic class names.

\begin{table}[t]
\centering
\caption{Surprisal scores for different navigation policies obtained during the agent exploration of the environment. 
}
\label{tab:curiosity}
\setlength{\tabcolsep}{.45em}
\begin{tabular}{lcc}
\toprule
\textbf{Navigation Module} &  &  \textbf{Surprisal} \\
\midrule
Random Exploration & & 0.333 \\
\midrule
\ours w/o Penalty for repeated actions (RGB only) & & 0.193 \\
\ours w/o Penalty for repeated actions (Depth only) & & 0.361 \\
\ours w/o Penalty for repeated actions (RGB + Depth) & & 0.439 \\
\midrule
\textbf{\ours} & & \textbf{0.697} \\
\bottomrule
\end{tabular}
\vspace{-0.1cm}
\end{table}

\begin{figure}[t]
\centering
\footnotesize
\setlength{\tabcolsep}{.3em}
\begin{tabular}{ccc:cc}
Random Exploration & w/o Penalty & & & \ours \\
\includegraphics[width=0.3\linewidth]{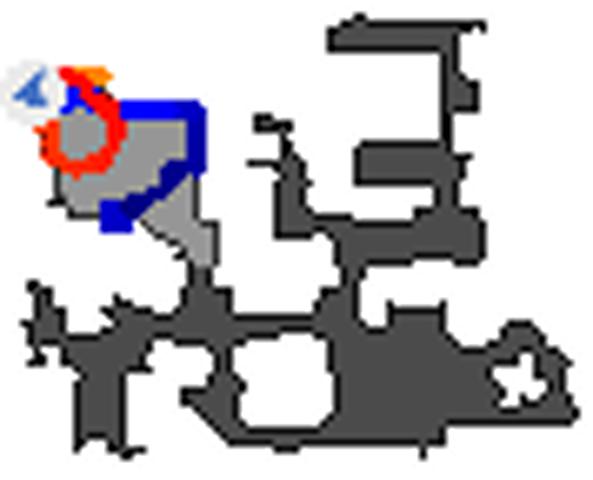} &
\includegraphics[width=0.3\linewidth]{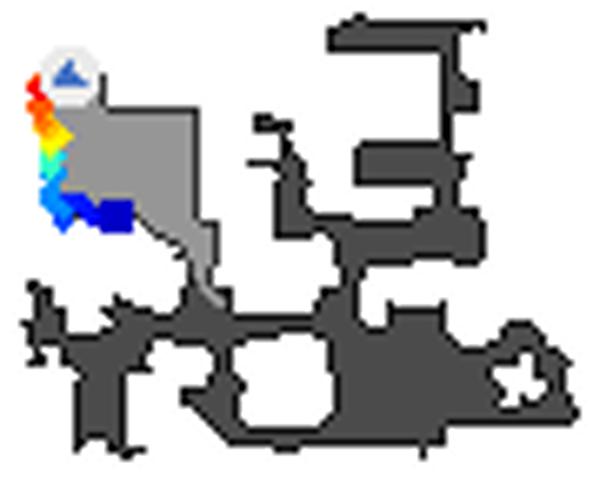} & & &
\includegraphics[width=0.3\linewidth]{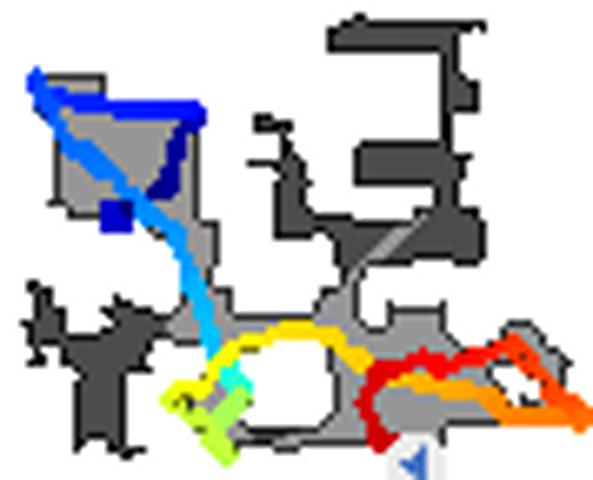} \\
\includegraphics[width=0.3\linewidth]{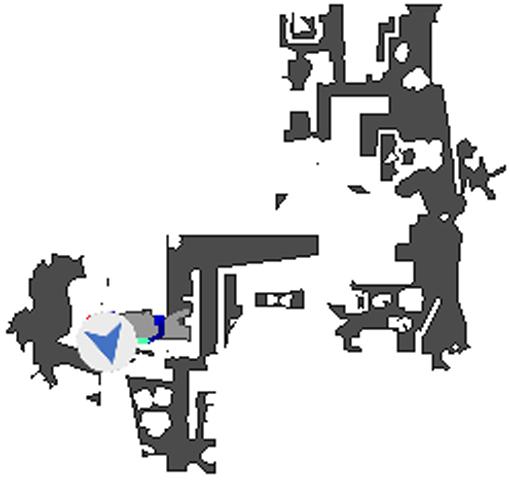} &
\includegraphics[width=0.3\linewidth]{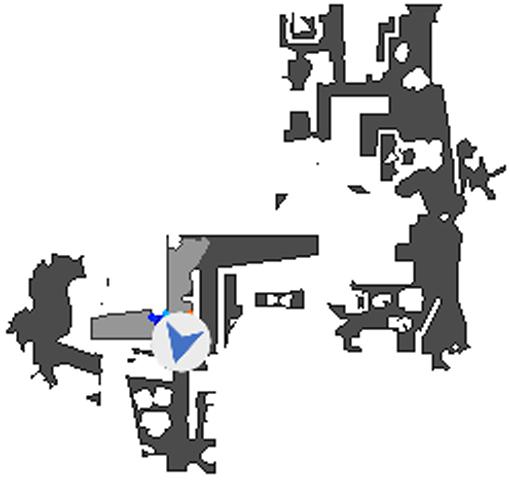} & & &
\includegraphics[width=0.3\linewidth]{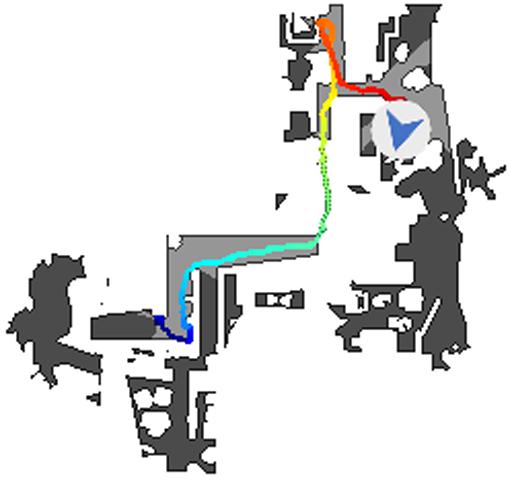} \\
\includegraphics[width=0.3\linewidth]{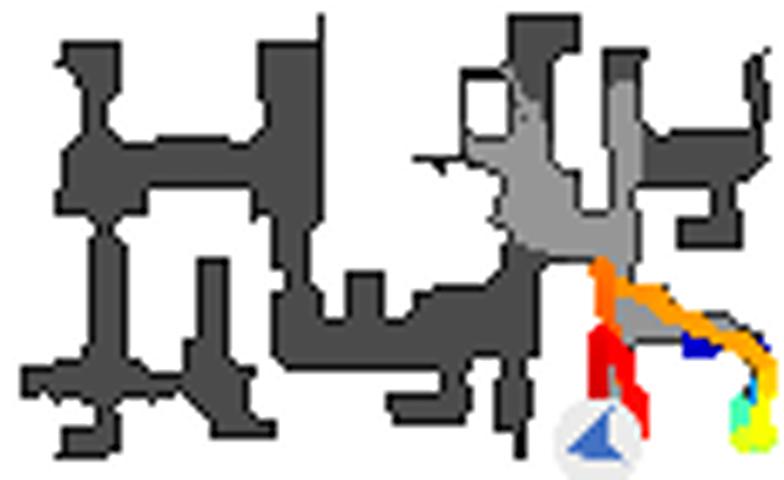} &
\includegraphics[width=0.3\linewidth]{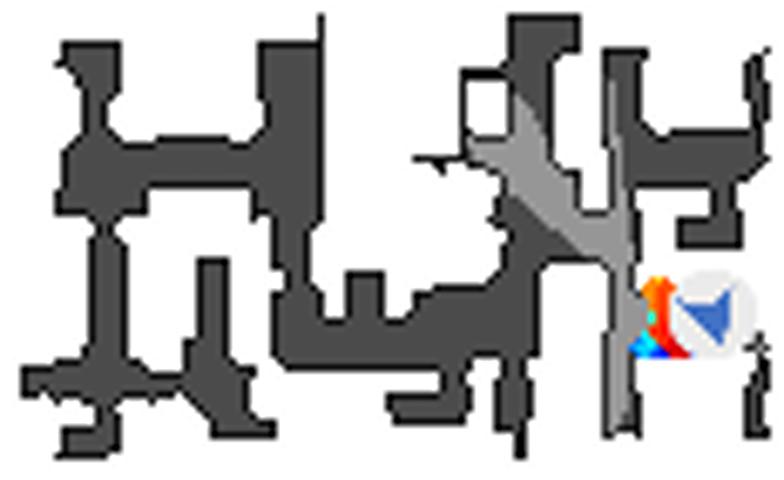} & & &
\includegraphics[width=0.3\linewidth]{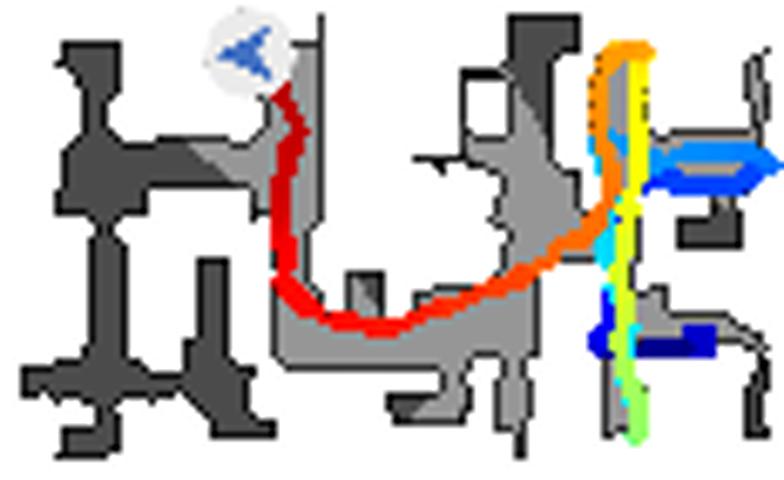} \\
\end{tabular}
\caption{Qualitative results of the agent trajectories in sample navigation episodes.}
\label{fig:navigation_maps}
\vspace{-0.15cm}
\end{figure}

\section{Experimental Results}

\subsection{Navigation results}
As defined in Sec.~\ref{sec:protocol}, we evaluate the performance of our navigation agents by computing the average surprisal score over test episodes.
Results are reported in Table~\ref{tab:curiosity} and show that our complete method (\ours) outperforms all other variants, achieving a significantly greater surprisal score than our method without penalty. In particular, the final performance greatly benefits from using both visual modalities (RGB and depth), instead of using a single visual modality to represent the scene.
Notably, random exploration (\eg~sampling $a_t$ from a uniform distribution over the available actions at each time step $t$) proves to be a strong baseline for this task, performing better than our single-modality RGB agent. Nonetheless, our final agent greatly outperforms the baselines, scoring $0.364$ and $0.258$ above the random policy and the vanilla curiosity-based agent respectively.

\begin{table*}[t]
\centering
\caption{Coverage and diversity results for different version of our captioning module. Results are reported for our three speaker policies using different thresholds to determine the agent's loquacity inside the episode.} 
\label{tab:captioning_results}
\setlength{\tabcolsep}{.35em}
\resizebox{\linewidth}{!}{
\begin{tabular}{lcccccccccccccccccc}
\toprule
& & \multicolumn{5}{c}{\textbf{Object-driven policy $\mathbf{(O\geq1)}$}} & & \multicolumn{5}{c}{\textbf{Object-driven policy $\mathbf{(O\geq3)}$}} & & \multicolumn{5}{c}{\textbf{Object-driven policy $\mathbf{(O\geq5)}$}}  \\
& & \multicolumn{5}{c}{\scriptsize$\mathsf{Loquacity=43.3}$} & & \multicolumn{5}{c}{\scriptsize$\mathsf{Loquacity=27.4}$} & & \multicolumn{5}{c}{\scriptsize$\mathsf{Loquacity=15.8}$}   \\
\cmidrule{3-7} \cmidrule{9-13} \cmidrule{15-19} 
\textbf{Captioning Module} & & $\mathsf{Cov}_{>1\%}$ & $\mathsf{Cov}_{>3\%}$ & $\mathsf{Cov}_{>5\%}$ & $\mathsf{Cov}_{>10\%}$ & $\mathsf{Div}$ & & $\mathsf{Cov}_{>1\%}$ & $\mathsf{Cov}_{>3\%}$ & $\mathsf{Cov}_{>5\%}$ & $\mathsf{Cov}_{>10\%}$ & $\mathsf{Div}$ & & $\mathsf{Cov}_{>1\%}$ & $\mathsf{Cov}_{>3\%}$ & $\mathsf{Cov}_{>5\%}$ & $\mathsf{Cov}_{>10\%}$ & $\mathsf{Div}$ \\
\midrule
\textbf{\ours}~(6 lay. as in~\cite{vaswani2017attention}) & & 0.456 & 0.550 & 0.609 & 0.706 & 0.386 & & 0.387 & 0.502 & 0.576 & 0.696 & 0.363 & & 0.348 & 0.468 & 0.549 & 0.691 & 0.352 \\
\textbf{\ours}~(3 lay.) & & 0.474 & 0.558 & 0.612 & 0.701 & 0.372 & & 0.384 & 0.497 & 0.571 & 0.691 & 0.350 & & 0.347 & 0.467 & 0.546 & 0.688 & 0.338 \\
\textbf{\ours}~(2 lay.) & & \textbf{0.485} & \textbf{0.579} & \textbf{0.637} & \textbf{0.727} & 0.368 & & \textbf{0.416} & \textbf{0.534} & \textbf{0.607} & \textbf{0.721} & 0.349 & & \textbf{0.373} & \textbf{0.497} & \textbf{0.577} & \textbf{0.713} & 0.340 \\
\textbf{\ours}~(1 lay.) & & 0.468 & 0.564 & 0.623 & 0.720 & \textbf{0.394} & & 0.400 & 0.519 & 0.593 & 0.713 & \textbf{0.377} & & 0.356 & 0.479 & 0.560 & 0.702 & \textbf{0.373} \\
\midrule
\midrule
 & & \multicolumn{5}{c}{\textbf{Depth-driven policy $\mathbf{(D>0.25)}$}} & & \multicolumn{5}{c}{\textbf{Depth-driven policy $\mathbf{(D>0.5)}$}} & & \multicolumn{5}{c}{\textbf{Depth-driven policy $\mathbf{(D>0.75)}$}}  \\
 & & \multicolumn{5}{c}{\scriptsize$\mathsf{Loquacity=38.5}$} & & \multicolumn{5}{c}{\scriptsize$\mathsf{Loquacity=31.1}$} & & \multicolumn{5}{c}{\scriptsize$\mathsf{Loquacity=14.8}$}   \\
\cmidrule{3-7} \cmidrule{9-13} \cmidrule{15-19} 
\textbf{Captioning Module} & & $\mathsf{Cov}_{>1\%}$ & $\mathsf{Cov}_{>3\%}$ & $\mathsf{Cov}_{>5\%}$ & $\mathsf{Cov}_{>10\%}$ & $\mathsf{Div}$ & & $\mathsf{Cov}_{>1\%}$ & $\mathsf{Cov}_{>3\%}$ & $\mathsf{Cov}_{>5\%}$ & $\mathsf{Cov}_{>10\%}$ & $\mathsf{Div}$ & & $\mathsf{Cov}_{>1\%}$ & $\mathsf{Cov}_{>3\%}$ & $\mathsf{Cov}_{>5\%}$ & $\mathsf{Cov}_{>10\%}$ & $\mathsf{Div}$ \\
\midrule
\textbf{\ours}~(6 lay. as in~\cite{vaswani2017attention}) & & 0.433 & 0.532 & 0.600 & 0.705 & 0.360 & & 0.420 & 0.519 & 0.585 & 0.701 & 0.346 & & 0.399 & 0.497 & 0.566 & 0.691 & 0.339 \\
\textbf{\ours}~(3 lay.) & & 0.427 & 0.524 & 0.588 & 0.700 & 0.349 & & 0.413 & 0.511 & 0.577 & 0.695 & 0.335 & & 0.394 & 0.491 & 0.559 & 0.685 & 0.330 \\
\textbf{\ours}~(2 lay.) & & \textbf{0.463} & \textbf{0.562} & \textbf{0.625} & \textbf{0.730} & 0.341 & & \textbf{0.449} & \textbf{0.550} & \textbf{0.612} & \textbf{0.726} & 0.330 & & \textbf{0.425} & \textbf{0.525} & \textbf{0.595} & \textbf{0.715} & 0.325 \\
\textbf{\ours}~(1 lay.) & & 0.448 & 0.548 & 0.613 & 0.723 & \textbf{0.371} & & 0.434 & 0.536 & 0.603 & 0.719 & \textbf{0.359} & & 0.412 & 0.513 & 0.583 & 0.708 & \textbf{0.355} \\
\midrule
\midrule
 & & \multicolumn{5}{c}{\textbf{Curiosity-driven policy $\mathbf{(S>0.7)}$}} & & \multicolumn{5}{c}{\textbf{Curiosity-driven policy $\mathbf{(S>0.85)}$}} & & \multicolumn{5}{c}{\textbf{Curiosity-driven policy $\mathbf{(S>1.0)}$}}  \\
 & & \multicolumn{5}{c}{\scriptsize$\mathsf{Loquacity=27.2}$} & & \multicolumn{5}{c}{\scriptsize$\mathsf{Loquacity=18.2}$} & & \multicolumn{5}{c}{\scriptsize$\mathsf{Loquacity=6.4}$}   \\
\cmidrule{3-7} \cmidrule{9-13} \cmidrule{15-19} 
\textbf{Captioning Module} & & $\mathsf{Cov}_{>1\%}$ & $\mathsf{Cov}_{>3\%}$ & $\mathsf{Cov}_{>5\%}$ & $\mathsf{Cov}_{>10\%}$ & $\mathsf{Div}$ & & $\mathsf{Cov}_{>1\%}$ & $\mathsf{Cov}_{>3\%}$ & $\mathsf{Cov}_{>5\%}$ & $\mathsf{Cov}_{>10\%}$ & $\mathsf{Div}$ & & $\mathsf{Cov}_{>1\%}$ & $\mathsf{Cov}_{>3\%}$ & $\mathsf{Cov}_{>5\%}$ & $\mathsf{Cov}_{>10\%}$ & $\mathsf{Div}$ \\
\midrule
\textbf{\ours}~(6 lay. as in~\cite{vaswani2017attention}) & & 0.425 & 0.523 & 0.588 & 0.703 & 0.356 & & 0.421 & 0.515 & 0.581 & 0.699 & 0.360 & & 0.422 & 0.518 & 0.583 & 0.702 & 0.364 \\
\textbf{\ours}~(3 lay.) & & 0.418 & 0.514 & 0.578 & 0.694 & 0.348 & & 0.413 & 0.506 & 0.571 & 0.691 & 0.350 & & 0.413 & 0.506 & 0.570 & 0.690 & 0.361 \\
\textbf{\ours}~(2 lay.) & & \textbf{0.453} & \textbf{0.552} & \textbf{0.617} & \textbf{0.726} & 0.340 & & \textbf{0.448} & \textbf{0.545} & \textbf{0.611} & \textbf{0.724} & 0.342 & & \textbf{0.448} & \textbf{0.545} & \textbf{0.610} & \textbf{0.723} & 0.349 \\
\textbf{\ours}~(1 lay.) & & 0.438 & 0.539 & 0.604 & 0.719 & \textbf{0.370} & & 0.433 & 0.530 & 0.597 & 0.716 & \textbf{0.373} & & 0.434 & 0.532 & 0.597 & 0.717 & \textbf{0.380} \\
\bottomrule
\end{tabular}
}
\vspace{-0.15cm}
\end{table*}

\tinytit{Qualitative Analysis}
In Fig.~\ref{fig:navigation_maps}, we report some top-down views from the testing scenes, together with the trajectory from three different navigation agents: the random baseline, our approach without the penalty for repeated action described in Sec.~\ref{sec:method_nav}, and our full model.
We notice that the agent without penalty usually remains in the starting area and thus has some difficulties in exploring the whole environment.
Instead, our complete model demonstrates better results as it is able to explore a much wider area within the environment.
Thus, we conclude that the addition of a penalty for repeated actions in the final reward function is of central importance when it comes to stimulating the agent towards the exploration of regions far from the starting point.

\subsection{Speaker Results}
Here, we provide quantitative and qualitative results for our speaker module, which is composed of a policy and a captioner. The policy is in charge of deciding when to activate the captioner, which in turns generates a description of the first-person view of the agent.
Results are reported in Table~\ref{tab:captioning_results} and discussed below.

\tinytit{Speaker Policy}
Among the three different policies, the object-driven speaker performs the best in terms of coverage and diversity. In particular, setting a low threshold ($\mathbf{O}\geq1$) provides the highest scores. At the same time, the agent tends to speak more often, which is desirable in a visually rich environment. As the threshold for $\mathbf{O}$ gets higher, performances get worse. This indicates that, as the number of object in the scene increases, there are many details that the captioner cannot describe. The same applies for the depth-driven policy: while the agent tends to describe well items that are closer, it experiences some troubles when facing an open space with more distant objects ($\mathbf{D} \geq 0.75$).

Instead, our curiosity-driven speaker shows a more peculiar behaviour: as the threshold grows, results get better in terms of diversity, while the coverage scores are quite stable (only $-0.005\%$ in terms of $\text{Cov}_{>1\%}$). 
It is also worth mentioning that our curiosity-based speaker can be adopted in any kind of environment, as the driving metric is computed from the raw RGB-D input. The same does not apply in an object-driven policy, since the agent needs semantic information.
Further, the curiosity-driven policy employs a learned metric, hence being more related to the exploration module.

From all these observations, we can conclude that curiosity not only helps training navigation agents, but also represents and important metric when bridging cross-modal components in embodied agents. 

\tinytit{Captioner}
When evaluating the captioning module, we compare the performance using a different number of encoding and decoding layers. As it can be seen from Table~\ref{tab:captioning_results}, the captioning model achieves the best results when composed of $2$ layers for coverage and $1$ layer for diversity. While this is in contrast with traditional Transformer-based models~\cite{vaswani2017attention}, that employ $6$ or more layers, it is in line with recent research on image captioning~\cite{cornia2020m2}, which finds beneficial to adopt fewer layers. At the same time, a more lightweight network can possibly be embedded in many embodied agents, thus being more appropriate for our task.

\begin{figure*}[t]
    \centering
    \includegraphics[width=\linewidth]{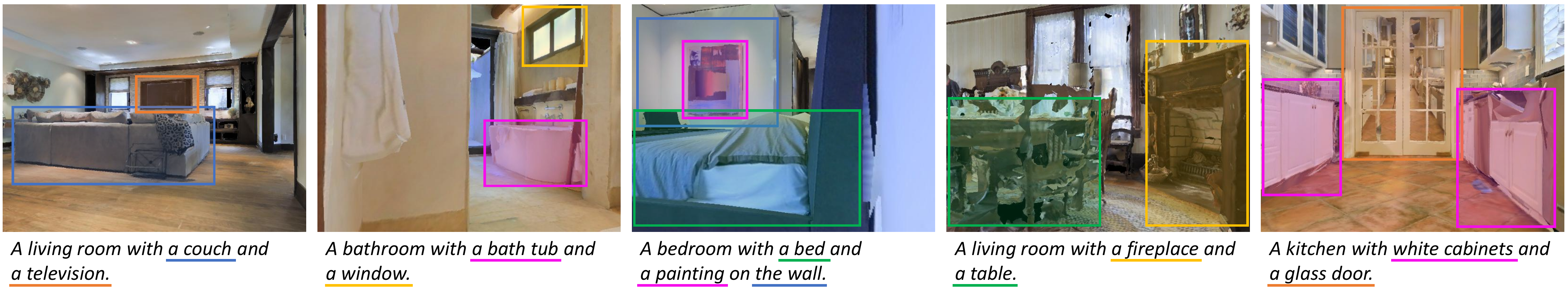}
    \caption{Sentences generated on sample images extracted from \ours navigation trajectories. For each image, we report the relevant objects present on the scene and we underline their mentions in the caption.}
    \label{fig:sample_captions}
    \vspace{-0.15cm}
\end{figure*}

\tinytit{Qualitative Analysis}
We report some qualitative results for \ours in Fig.~\ref{fig:sample_captions}. To ease visualization, we underline the items mentioned by the captioner in the sentence, and highlight them with a bounding box of the same color in the corresponding input image. Our agent can explain the scene perceived from a first-person, egocentric point of view. We can notice that \ours identifies all the main objects in the environment and produces a suitable description even when the view is partially occluded.

%% file: 05-conclusion.tex
\section{Conclusion}
\label{sec:conclusion}
%

In this work, we have presented a new setting for embodied AI that is composed of two tasks: exploration and captioning. Our agent \ours uses intrinsic rewards applied to navigation in a photorealistic environment and a novel speaker module that generates captions. The captioner produces sentences according to a speaker policy that could be based on three metrics: object-driven, depth-driven, and curiosity-driven. The experiments show that \ours is able to generalize to unseen environments in terms of exploration, while the speaker policy functions to filter the number of time steps where the caption is actually generated.
We hope that our work serves as a starting point for future research on this new coupled-task of exploration and captioning. Our results with curiosity-based navigation in photorealistic environments and with the speaker module motivate further works in this direction.

%% file: main.bbl
\begin{thebibliography}{10}
\providecommand{\url}[1]{#1}
\csname url@samestyle\endcsname
\providecommand{\newblock}{\relax}
\providecommand{\bibinfo}[2]{#2}
\providecommand{\BIBentrySTDinterwordspacing}{\spaceskip=0pt\relax}
\providecommand{\BIBentryALTinterwordstretchfactor}{4}
\providecommand{\BIBentryALTinterwordspacing}{\spaceskip=\fontdimen2\font plus
\BIBentryALTinterwordstretchfactor\fontdimen3\font minus
  \fontdimen4\font\relax}
\providecommand{\BIBforeignlanguage}[2]{{%
\expandafter\ifx\csname l@#1\endcsname\relax
\typeout{** WARNING: IEEEtran.bst: No hyphenation pattern has been}%
\typeout{** loaded for the language `#1'. Using the pattern for}%
\typeout{** the default language instead.}%
\else
\language=\csname l@#1\endcsname
\fi
#2}}
\providecommand{\BIBdecl}{\relax}
\BIBdecl

\bibitem{anderson2018vision}
P.~Anderson, Q.~Wu, D.~Teney, J.~Bruce, M.~Johnson, N.~S{\"u}nderhauf, I.~Reid,
  S.~Gould, and A.~van~den Hengel, ``Vision-and-language navigation:
  Interpreting visually-grounded navigation instructions in real
  environments,'' in \emph{CVPR}, 2018.

\bibitem{Savva_2019_ICCV}
M.~Savva, A.~Kadian, O.~Maksymets, Y.~Zhao, E.~Wijmans, B.~Jain, J.~Straub,
  J.~Liu, V.~Koltun, J.~Malik, D.~Parikh, and D.~Batra, ``{Habitat: A Platform
  for Embodied AI Research},'' in \emph{ICCV}, 2019.

\bibitem{xia2018gibson}
F.~Xia, A.~R. Zamir, Z.~He, A.~Sax, J.~Malik, and S.~Savarese, ``Gibson env:
  Real-world perception for embodied agents,'' in \emph{CVPR}, 2018.

\bibitem{karpathy2015deep}
A.~Karpathy and L.~Fei-Fei, ``Deep visual-semantic alignments for generating
  image descriptions,'' in \emph{CVPR}, 2015.

\bibitem{anderson2018bottom}
P.~Anderson, X.~He, C.~Buehler, D.~Teney, M.~Johnson, S.~Gould, and L.~Zhang,
  ``Bottom-up and top-down attention for image captioning and visual question
  answering,'' in \emph{CVPR}, 2018.

\bibitem{cornia2020m2}
M.~Cornia, M.~Stefanini, L.~Baraldi, and R.~Cucchiara, ``{Meshed-Memory
  Transformer for Image Captioning},'' in \emph{CVPR}, 2020.

\bibitem{wijmans2019dd}
E.~Wijmans, A.~Kadian, A.~Morcos, S.~Lee, I.~Essa, D.~Parikh, M.~Savva, and
  D.~Batra, ``{DD-PPO: Learning Near-Perfect PointGoal Navigators from 2.5
  Billion Frames},'' in \emph{ICLR}, 2020.

\bibitem{agrawal2016learning}
P.~Agrawal, A.~V. Nair, P.~Abbeel, J.~Malik, and S.~Levine, ``Learning to poke
  by poking: Experiential learning of intuitive physics,'' in \emph{NeurIPS},
  2016.

\bibitem{pathak2017curiosity}
D.~Pathak, P.~Agrawal, A.~A. Efros, and T.~Darrell, ``Curiosity-driven
  exploration by self-supervised prediction,'' in \emph{CVPR Workshops}, 2017.

\bibitem{burda2018large}
Y.~Burda, H.~Edwards, D.~Pathak, A.~Storkey, T.~Darrell, and A.~A. Efros,
  ``Large-scale study of curiosity-driven learning,'' \emph{arXiv preprint
  arXiv:1808.04355}, 2018.

\bibitem{ramakrishnan2020exploration}
S.~K. Ramakrishnan, D.~Jayaraman, and K.~Grauman, ``An exploration of embodied
  visual exploration,'' \emph{arXiv preprint arXiv:2001.02192}, 2020.

\bibitem{cornia2019smart}
M.~Cornia, L.~Baraldi, and R.~Cucchiara, ``{SMArT: Training Shallow
  Memory-aware Transformers for Robotic Explainability},'' in \emph{ICRA},
  2020.

\bibitem{anjomshoae2019explainable}
S.~Anjomshoae, A.~Najjar, D.~Calvaresi, and K.~Fr{\"a}mling, ``Explainable
  agents and robots: Results from a systematic literature review,'' in
  \emph{AAMAS}, 2019.

\bibitem{vaswani2017attention}
A.~Vaswani, N.~Shazeer, N.~Parmar, J.~Uszkoreit, L.~Jones, A.~N. Gomez,
  {\L}.~Kaiser, and I.~Polosukhin, ``Attention is all you need,'' in
  \emph{NeurIPS}, 2017.

\bibitem{anderson2018evaluation}
P.~Anderson, A.~Chang, D.~S. Chaplot, A.~Dosovitskiy, S.~Gupta, V.~Koltun,
  J.~Kosecka, J.~Malik, R.~Mottaghi, M.~Savva \emph{et~al.}, ``On evaluation of
  embodied navigation agents,'' \emph{arXiv preprint arXiv:1807.06757}, 2018.

\bibitem{zhu2017target}
Y.~Zhu, R.~Mottaghi, E.~Kolve, J.~J. Lim, A.~Gupta, L.~Fei-Fei, and A.~Farhadi,
  ``Target-driven visual navigation in indoor scenes using deep reinforcement
  learning,'' in \emph{ICRA}, 2017.

\bibitem{oudeyer2009intrinsic}
P.-Y. Oudeyer and F.~Kaplan, ``{What is intrinsic motivation? A typology of
  computational approaches},'' \emph{Frontiers in Neurorobotics}, vol.~1, p.~6,
  2009.

\bibitem{schmidhuber2010formal}
J.~Schmidhuber, ``{Formal Theory of Creativity, Fun, and Intrinsic
  Motivation},'' \emph{IEEE Trans. on Autonomous Mental Development}, vol.~2,
  no.~3, pp. 230--247, 2010.

\bibitem{sun2011planning}
Y.~Sun, F.~Gomez, and J.~Schmidhuber, ``Planning to be surprised: Optimal
  bayesian exploration in dynamic environments,'' in \emph{AGI}, 2011.

\bibitem{klyubin2005empowerment}
A.~S. Klyubin, D.~Polani, and C.~L. Nehaniv, ``Empowerment: A universal
  agent-centric measure of control,'' in \emph{CEC}, 2005.

\bibitem{mohamed2015variational}
S.~Mohamed and D.~J. Rezende, ``Variational information maximisation for
  intrinsically motivated reinforcement learning,'' in \emph{NeurIPS}, 2015.

\bibitem{houthooft2016vime}
R.~Houthooft, X.~Chen, Y.~Duan, J.~Schulman, F.~De~Turck, and P.~Abbeel,
  ``{VIME: Variational Information Maximizing Exploration},'' in
  \emph{NeurIPS}, 2016.

\bibitem{bellemare2016unifying}
M.~Bellemare, S.~Srinivasan, G.~Ostrovski, T.~Schaul, D.~Saxton, and R.~Munos,
  ``Unifying count-based exploration and intrinsic motivation,'' in
  \emph{NeurIPS}, 2016.

\bibitem{tang2017exploration}
H.~Tang, R.~Houthooft, D.~Foote, A.~Stooke, O.~X. Chen, Y.~Duan, J.~Schulman,
  F.~DeTurck, and P.~Abbeel, ``{\#Exploration: A study of count-based
  exploration for deep reinforcement learning},'' in \emph{NeurIPS}, 2017.

\bibitem{bellemare2013arcade}
M.~G. Bellemare, Y.~Naddaf, J.~Veness, and M.~Bowling, ``The arcade learning
  environment: An evaluation platform for general agents,'' \emph{J. of
  Artificial Intelligence Research}, vol.~47, pp. 253--279, 2013.

\bibitem{brockman2016openai}
G.~Brockman, V.~Cheung, L.~Pettersson, J.~Schneider, J.~Schulman, J.~Tang, and
  W.~Zaremba, ``{OpenAI Gym},'' \emph{arXiv preprint arXiv:1606.01540}, 2016.

\bibitem{kempka2016vizdoom}
M.~Kempka, M.~Wydmuch, G.~Runc, J.~Toczek, and W.~Ja{\'s}kowski, ``{ViZDoom: A
  doom-based ai research platform for visual reinforcement learning},'' in
  \emph{IEEE Conf. CIG}, 2016.

\bibitem{beattie2016deepmind}
C.~Beattie, J.~Z. Leibo, D.~Teplyashin, T.~Ward, M.~Wainwright, H.~K{\"u}ttler,
  A.~Lefrancq, S.~Green, V.~Vald{\'e}s, A.~Sadik \emph{et~al.}, ``{DeepMind
  Lab},'' \emph{arXiv preprint arXiv:1612.03801}, 2016.

\bibitem{xu2015show}
K.~Xu, J.~Ba, R.~Kiros, K.~Cho, A.~Courville, R.~Salakhudinov, R.~Zemel, and
  Y.~Bengio, ``Show, attend and tell: Neural image caption generation with
  visual attention,'' in \emph{ICML}, 2015.

\bibitem{rennie2017self}
S.~J. Rennie, E.~Marcheret, Y.~Mroueh, J.~Ross, and V.~Goel, ``Self-critical
  sequence training for image captioning,'' in \emph{CVPR}, 2017.

\bibitem{vinyals2017show}
O.~Vinyals, A.~Toshev, S.~Bengio, and D.~Erhan, ``{Show and Tell: Lessons
  Learned from the 2015 MSCOCO Image Captioning Challenge},'' \emph{IEEE Trans.
  PAMI}, vol.~39, no.~4, pp. 652--663, 2017.

\bibitem{herdade2019image}
S.~Herdade, A.~Kappeler, K.~Boakye, and J.~Soares, ``{Image Captioning:
  Transforming Objects into Words},'' in \emph{NeurIPS}, 2019.

\bibitem{achiam2017surprise}
J.~Achiam and S.~Sastry, ``Surprise-based intrinsic motivation for deep
  reinforcement learning,'' in \emph{NeurIPS Workshops}, 2017.

\bibitem{ren2017faster}
S.~Ren, K.~He, R.~Girshick, and J.~Sun, ``{Faster R-CNN: towards real-time
  object detection with region proposal networks},'' \emph{IEEE Trans. PAMI},
  vol.~39, no.~6, pp. 1137--1149, 2017.

\bibitem{Matterport3D}
A.~Chang, A.~Dai, T.~Funkhouser, M.~Halber, M.~Niessner, M.~Savva, S.~Song,
  A.~Zeng, and Y.~Zhang, ``{Matterport3D: Learning from RGB-D Data in Indoor
  Environments},'' in \emph{3D Vision}, 2017.

\bibitem{kuhn1955hungarian}
H.~W. Kuhn, ``{The Hungarian method for the assignment problem},'' \emph{Naval
  Research Logistics Quarterly}, vol.~2, no. 1-2, pp. 83--97, 1955.

\bibitem{schulman2017proximal}
J.~Schulman, F.~Wolski, P.~Dhariwal, A.~Radford, and O.~Klimov, ``Proximal
  policy optimization algorithms,'' \emph{arXiv preprint arXiv:1707.06347},
  2017.

\bibitem{kingma2015adam}
D.~Kingma and J.~Ba, ``Adam: a method for stochastic optimization,'' in
  \emph{ICLR}, 2015.

\bibitem{krishnavisualgenome}
R.~Krishna, Y.~Zhu, O.~Groth, J.~Johnson, K.~Hata, J.~Kravitz, S.~Chen,
  Y.~Kalantidis, L.-J. Li, D.~A. Shamma, M.~Bernstein, and L.~Fei-Fei,
  ``{Visual Genome: Connecting Language and Vision Using Crowdsourced Dense
  Image Annotations},'' \emph{IJCV}, vol. 123, no.~1, pp. 32--73, 2017.

\bibitem{lin2014microsoft}
T.-Y. Lin, M.~Maire, S.~Belongie, J.~Hays, P.~Perona, D.~Ramanan,
  P.~Doll{\'a}r, and C.~L. Zitnick, ``{Microsoft COCO: Common Objects in
  Context},'' in \emph{ECCV}, 2014.

\bibitem{vedantam2015cider}
R.~Vedantam, C.~Lawrence~Zitnick, and D.~Parikh, ``{CIDEr: Consensus-based
  Image Description Evaluation},'' in \emph{CVPR}, 2015.

\bibitem{pennington2014glove}
J.~Pennington, R.~Socher, and C.~D. Manning, ``{GloVe: Global Vectors for Word
  Representation},'' in \emph{EMNLP}, 2014.

\end{thebibliography}
